\documentclass[sigconf,natbib=true,anonymous=false]{acmart}

\AtBeginDocument{%
  \providecommand\BibTeX{{%
    \normalfont B\kern-0.5em{\scshape i\kern-0.25em b}\kern-0.8em\TeX}}}

\copyrightyear{2023}
\acmYear{2023}
\setcopyright{acmlicensed}\acmConference[SIGIR '23]{Proceedings of the 46th International ACM SIGIR Conference on Research and Development in Information Retrieval}{July 23--27, 2023}{Taipei, Taiwan}
\acmBooktitle{Proceedings of the 46th International ACM SIGIR Conference on Research and Development in Information Retrieval (SIGIR '23), July 23--27, 2023, Taipei, Taiwan}
\acmPrice{15.00}
\acmDOI{10.1145/3539618.3591877}
\acmISBN{978-1-4503-9408-6/23/07}

\usepackage{xspace}
\usepackage{tabularx}
\usepackage{multicol}
\usepackage{multirow}
\usepackage{adjustbox}
\usepackage{subfigure}
\usepackage{soul}

\usepackage{amsthm,amsmath,amsfonts,bm,xspace}
\usepackage{color}

\definecolor{cerulean}{rgb}{0.0,0.48,0.65}
\definecolor{caribbean}{rgb}{0.0,0.8,0.6}
\definecolor{paleblue}{rgb}{0.82,0.93,1}
\definecolor{auburn}{rgb}{0.43,0.21,0.1}
\definecolor{byzantium}{rgb}{0.44,0.16,0.39}
\definecolor{byzantine}{rgb}{0.74,0.2,0.64}
\definecolor{violet}{rgb}{0.54,0.17,0.89}
\definecolor{cornell}{rgb}{0.7,0.11,0.11}
\definecolor{lgray}{gray}{0.85}
\definecolor{mgray}{gray}{0.7}
\definecolor{dgreen}{rgb}{0,0.55,0}
\definecolor{mgreen}{rgb}{0,0.7,0}
\definecolor{palered}{rgb}{1,0.70,0.75}
\definecolor{pastelyellow}{rgb}{0.99,0.99,0.65}
\definecolor{bleudefrance}{rgb}{0.19, 0.55, 0.91}
\definecolor{carminepink}{rgb}{0.92, 0.3, 0.26}

\newcommand{\system}[1]{\textsc{#1}}
\newcommand{\data}[1]{\textsc{#1}}

\newcommand{\bert}{\system{Zh-BERT}\xspace}
\newcommand{\roberta}{\system{Zh-RoBERTa}\xspace}

\newcommand{\gpt}{\system{GPT3}\xspace}

\newcommand{\ourcorpus}{\data{SocialDial}\xspace}

\newcommand{\moralInt}{\data{Moral Integrity}\xspace}
\newcommand{\PersuasionforGood}{\data{PersuasionForGood}\xspace}
\newcommand{\FactAct}{\data{FactAct}\xspace}

\newcommand\redsout{\bgroup\markoverwith{\textcolor{red}{\rule[0.5ex]{2pt}{0.8pt}}}\ULon}

\newcommand{\eg}{e.g.,}
\newcommand{\ie}{i.e.,}

\newcommand\code[1]{ {\small{\url{#1}}}}

\def\eqref#1{(\ref{#1})}

\def\1{\bm{1}}

\def\rmX{{\mathbf{X}}}

\def\vx{{\bm{x}}}

\DeclareMathAlphabet{\mathsfit}{\encodingdefault}{\sfdefault}{m}{sl}
\SetMathAlphabet{\mathsfit}{bold}{\encodingdefault}{\sfdefault}{bx}{n}

\begin{document}

\title{\ourcorpus: A Benchmark for Socially-Aware Dialogue Systems}




\newcommand{\heart}{\ensuremath\heartsuit}
\newcommand{\solidheart}{\ensuremath\varheartsuit}
\newcommand{\butt}{\rotatebox[origin=c]{180}{\solidheart}}

\author{Haolan Zhan\textsuperscript{\heart},\quad Zhuang Li\textsuperscript{\heart},\quad Yufei Wang\textsuperscript{\heart}, \quad Linhao Luo\textsuperscript{\heart}, \quad Tao Feng\textsuperscript{\heart}, \quad Xiaoxi Kang\textsuperscript{\ddag}, \\
\quad Yuncheng Hua\textsuperscript{\heart}, \quad Lizhen Qu\textsuperscript{\heart}*, \quad Lay-Ki Soon\textsuperscript{\ddag}, \quad 
Suraj Sharma\textsuperscript{\butt}, \\ \quad Ingrid Zukerman\textsuperscript{\heart}, \quad  Zhaleh Semnani-Azad\textsuperscript{\butt},
\quad Gholamreza Haffari\textsuperscript{\heart}*}

\makeatletter
\def\authornotetext#1{
\if@ACM@anonymous\else
    \g@addto@macro\@authornotes{
    \stepcounter{footnote}\footnotetext{#1}}
\fi}
\makeatother
\authornotetext{Corresponding authors.}

\affiliation{
 \institution{\textsuperscript{\rm \heart}Department of Data Science and Artificial Intelligence, Monash University, Australia} \country{}
\institution{\textsuperscript{\rm \ddag}School of Information Technology, Monash University, Malaysia}  
 \institution{\textsuperscript{\rm \butt}California State University, Northridge, Unitied States}
 }
 \email{
 {firstname.lastname}@monash.edu, 
 {yufei.wang1, devin.hua, Soon.LayKi}@monash.edu}
 
 \email{{suraj.sharma, 
  zhaleh.semnaniazad}@csun.edu
 }

 \def\authors{Haolan Zhan, Zhuang Li, Yufei Wang, Linhao Luo, Tao Feng,  Xiaoxi Kang, Yuncheng Hua, Lizhen Qu, Lay-Ki Soon, Suraj Sharma, Ingrid Zukerman, Zhaleh Semnani-Azad, Gholamreza Haffari}

\renewcommand{\shortauthors}{Zhan and Li, et al.}

\begin{abstract}
   \textit{\textbf{Content Warning}: this paper may contain content that is offensive or upsetting.}

Dialogue systems have been widely applied in many scenarios and are now more powerful and ubiquitous than ever before. With large neural models and massive available data, current dialogue systems have access to more knowledge than any people in their life. However, current dialogue systems still do not perform at a human level. One major gap between conversational agents and  humans lies in their abilities to be aware of \textit{social norms}. The development of socially-aware dialogue systems is impeded due to the lack of resources. In this paper, we present the first socially-aware dialogue corpus -- \ourcorpus, based on Chinese social culture. \ourcorpus consists of two parts: 1,563 multi-turn dialogues between two human speakers with fine-grained labels, and 4,870  synthetic conversations generated by ChatGPT. The human corpus covers five categories of social norms, which have 14 sub-categories in total. Specifically, it contains  social factor annotations including \emph{social relation}, \emph{context}, \emph{social distance}, and \emph{social norms}.  However, collecting sufficient socially-aware dialogues is costly. Thus, we harness the power of ChatGPT and devise an ontology-based synthetic data generation framework. This framework is able to generate synthetic data at scale. To ensure the quality of synthetic dialogues, we design several mechanisms for quality control during data collection. Finally, we evaluate our dataset using several pre-trained models, such as BERT and RoBERTa. Comprehensive empirical results based on state-of-the-art neural models demonstrate that modeling of social norms for dialogue systems is a promising research direction.
To the best of our knowledge, \ourcorpus is the first socially-aware dialogue dataset that covers multiple social factors and has fine-grained labels.\footnote{Dataset and baselines are shared at: \url{https://github.com/zhanhl316/SocialDial}.}

\end{abstract}







\maketitle

\section{Introduction}

\begin{figure}
    \centering  
    \includegraphics[width=0.42\textwidth]{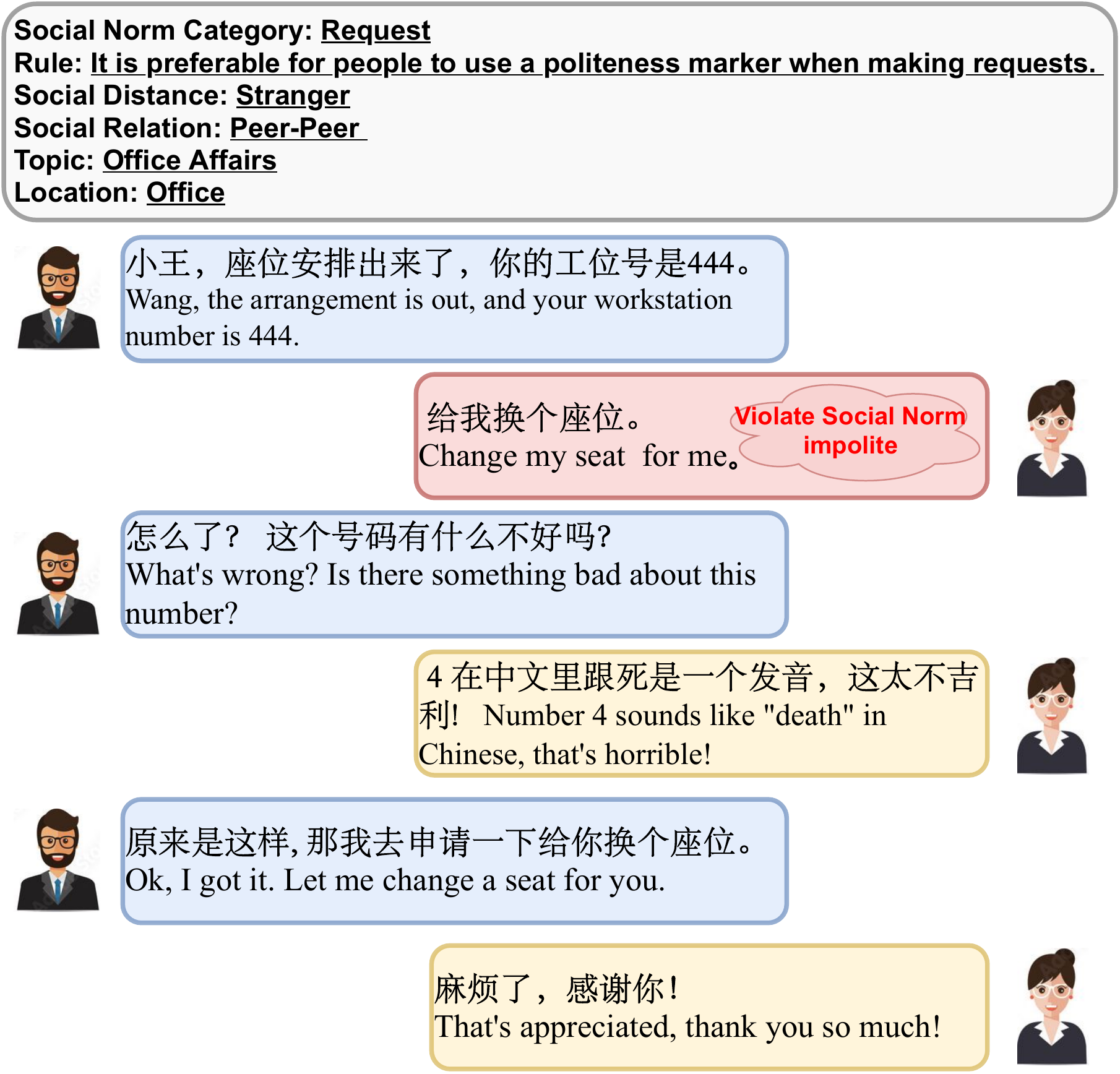}
    \caption{A demonstration of socially-aware dialogue in \ourcorpus. Each dialogue starts with a social knowledge profile (in the grey box). In the dialogue, each turn is annotated with a tag whether it is in violation of specific social norms. }
    \label{fig:intro}
\vspace{-1mm}
\end{figure}

With a population of over 1.4 billion, China is the most populous country in the world and the second largest economy. Hence, it is important to understand Chinese social norms when interacting with Chinese people~\cite{manion1993retirement,yang1993chinese,chen2019social,xiao2020social}. Social norms in a culture refer to acceptable behaviours and a shared understanding about what people should or should not do in that culture~\cite{hovy-yang-2021-importance}. In this work, we are particularly interested in social norms in conversations, as demonstrated by the example in Figure~\ref{fig:intro}. In a conversation between two colleagues, it is not polite for a colleague (Wang) to make a direct request without using any politeness markers. However, this impolite behavior may have been triggered by her interlocutor's cultural lack of awareness that the number four sounds like ``death'' in Chinese. Violation of social norms can potentially lead to unexpected damage, such as disengagement and break of relationships. 

Norms in conversations are often strongly associated with other supporting social factors, such as social relations and social context~\cite{capurucco2009building}. Different social factors often lead to different acceptable or unacceptable behaviors in a conversation~\cite{ziems2022moralIntegrityCorpus,hovy-yang-2021-importance}. 
For instance, if the two interlocutors are family members, it is culturally appropriate to make direct requests even without a politeness marker. 
Despite the importance of understanding the influence of social factors on social norms, to the best of our knowledge, there is no dialogue corpus annotated with rich social factors for studying Chinese social norms in conversations.


To address this research gap, we construct the \textit{first} dialogue benchmark, \ourcorpus, for investigating Chinese social norms. As prior studies in social science ~\cite{walton1991behavioral,olekalns2003testing,leung2011within,blader2012differentiating} already shed light on what are possible social factors influencing Chinese social norms, we start by building an ontology of social factors grounded in politeness theory~\cite{mao1994beyond}. Based on this ontology, we explore two low-cost approaches to collect dialogues annotated with social factors. The first approach is to ask crowd-workers to write dialogues according to the social norms in the ontology. We collect 1,563 dialogues in total as human-written dialogues. Since this corpus is relatively small, we automatically generate over 4,870 synthetic dialogues annotated with social factors from the ontology by leveraging ChatGPT\footnote{https://chat.openai.com/chat}, which is well known for its ability to generate high-quality natural texts. To assess the quality of those dialogues, we conduct a comprehensive human evaluation to rate the collected dialogues with respect to a variety of criteria. In addition, we conduct extensive experiments to evaluate the predictability of social factors from dialogues (when they are latent), using state-of-the-art (SOTA) models including GPT-3~\cite{NEURIPS2020_1457c0d6}. We also demonstrate the usefulness of supporting social factors by incorporating them into social norm violation detection.
Our key  findings are:
\begin{itemize}
    \item Most of synthetic dialogues generated by ChatGPT are considered as natural and coherent as those written by crowd-workers. Many of the automatically generated annotations even exhibit slightly higher quality than those produced by humans according to our studies. 
    \item The supervised SOTA models for social factor prediction fail to generalize well on human-written dialogues, if those models are only trained on synthetic dialogues. However, the synthetic data can be used to boost the performance of all those models if the models are trained on a combination of synthetic and human-written dialogues. Thus, \ourcorpus is a valuable resource to study how to effectively use synthetic data for dialogue research.
    \item The supporting social factors contribute significantly to social norm violation detection by more than double the F1 scores of GPT-3. Therefore, this benchmark facilitates research on investigating social factors for socially-aware NLP applications.
\end{itemize}

\section{Related work}
\label{sec:relatd_work}
Previous research related to this paper can be broadly categorized into \emph{Social Implications in Free Text}, \emph{Social Context in Dialogues}, \emph{Synthetic Dialogue Generation} and \emph{Social Factors from Social Science Perspective}, which we cover in the rest of this section.


\subsection{Social Implications in Free Text}
This line of research aims to express  social implications of common events 
through natural language text. \citet{rashkin-etal-2018-event2mind} study 
human intentions and reactions from event phrases, \eg\ ``X drops a hint'' 
indicates that X's intention is to be subtle. 
\citet{DBLP:conf/aaai/SapBABLRRSC19} propose a benchmark with 877k items of commonsense knowledge using nine different ``If-Then'' relations. They show that neural models trained on this benchmark gain a commonsense reasoning capability. \citet{forbes-etal-2020-social} propose \emph{Social-Chemistry-101}, a large-scale corpus that catalogs 292k rules-of-thumb such as ``It is rude to run a blender at 5am''  as basic conceptual units.
\citet{sap-etal-2020-social} introduces the \emph{Social Bias Frames} benchmark, which is a hybrid of free-text and categorical features to reason about potential social bias implications in natural language.
\citet{tay-etal-2020-rather} proposes a novel benchmark on predicting preferable options
from two sentences describing scenarios that may involve social and cultural situations.
\citet{acharya2020towards} incorporate cultural differences in commonsense reasoning 
tasks by presenting several fairly universal rituals, such as birth, marriage, new year and birthdays, to annotators from the U.S. and India.
\citet{hendrycks2021aligning} introduce the ETHICS dataset -- a new benchmark that spans 
concepts in justice, well-being, duties, virtues and commonsense morality, where the 
task is to predict widespread moral judgments about diverse text scenarios.
Finally, \citet{DBLP:conf/aaai/LourieBC21} propose the large-scale SCRUPLES dataset, which
includes 625,000 ethical judgments over 32,000 real-life anecdotes. 
The main difference between \ourcorpus and these research efforts is that \ourcorpus sheds light on social factor analysis in dialogues with multi-turn interactions, 
which is more challenging than in a single plain text.
{Furthermore, \ourcorpus enables the study of the impact of social factors on norm adherence and violation in multi-turn dialogues.}


\subsection{Social Context in Dialogues}
This line of research focuses on extracting or incorporating social context in dialogues.
\citet{zhang-etal-2018-personalizing} propose to integrate speaker personas, expressed 
with several sentences (\eg\ ``I am a vegetarian''), into  chit-chat dialogue agents to produce responses with consistent personal information
(\eg\ the speaker is always a vegetarian across the long conversation). They propose to use memory-based neural networks to store this information and
successfully outperform models without persona profiles. \citet{chawla2021towards,chawla2022opponent}
focus on extracting emotion trajectories and objectives from various types of dialogues. \citet{DBLP:conf/acl/LiuZDSLYJH20} propose a novel Emotional Support Conversation benchmark to incorporate emotional support into dialogue systems. They provide rich annotations in
a help-seeker and supporter mode with comprehensive data quality controls. \citet{fung2022normsage} extract social norm rules from large-scale dialogue corpora using large-scale pre-trained language models. 
{However, each of these research efforts only utilizes a particular subset of social factors in the dialogues, while \ourcorpus is built on top of a comprehensive social knowledge ontology (Section~\ref{sec:ontology}). }

\subsection{Synthetic Data for Dialogues}
{Synthetic data are commonly used as data augmentation for low-resource dialogue systems.} \citet{lee-etal-2022-gpt} use  GPT-3  to generate synthetic empathetic dialogues through 
prompt-based in-context learning in both zero-shot and few-shot settings. To improve generation quality, 
they propose novel in-context example selection methods that utilize emotion and situational information.
Similarly, \citet{chen2023places} use LLMs to generate synthetic social dialogues covering various daily 
topics, such as school life and relationship. Their human and automatic evaluations show that the quality of such dialogues is similar to that 
of dialogues collected from people.
\citet{DBLP:conf/icml/DaiCZARGG22} propose a novel task called \emph{Dialogue Inpainting}, which transforms 
input documents into a two-party conversation between the document writer and a virtual reader, who asks
questions between the writer’s utterances. 
Human evaluation shows that the resulting synthetic dialogues are of high quality, and substantially 
improve the performance of SOTA conversational question-answering systems.
{Compared with existing methods, the synthetic dialogues in \ourcorpus are generated using ChatGPT, via a novel ontology-based framework, which enables ChatGPT to generate diverse dialogues at scale by combining different sets of symbolic social factors.}

\subsection{Social Factors from Social Science Perspective}
Previous research in social science have demonstrated the relation between different social factors. The politeness assessment formula in \citet{brown1987politeness}'s politeness theory shows that the power and socio-psychological distances between participants in a dialogue obey a role in the cognitive assessment of what behaviors are deemed appropriate in a situation. Other research has found that social norms differ between individuals of different levels of closeness and power distance~\cite{van2011breaking,oetzel2001face}.
Additionally, recent research in social science has emphasized the need to consider the context
when understanding social norms~\cite{van2019dynamic,figueiredo2019cultural,sarathy2017learning}. Different cultures and locations are associated with different norms, differ in their sensitivity to context, and differ in the importance placed on norms~\cite{gelfand2011differences,hall1976beyond}. These research efforts motivate us to build the social knowledge ontology described in Section~\ref{sec:ontology}.



\section{The \ourcorpus Benchmark}
\label{sec:corpus}
In order to study social norms in conversations and build models to help people overcome socio-cultural misunderstandings in conversations, in this section, we present the first benchmark for Chinese culture norm. 
Rooted in politeness theory~\cite{mao1994beyond}, it includes an ontology of social factors relevant to social norms in Chinese culture, 
a crowd-sourced corpus annotated with the social factors from the ontology, and a synthetic dialogue corpus containing social factor annotations generated by ChatGPT. Our human evaluations on crowd-sourced and synthetic corpora show that i)~the dialogue quality of crowd-sourced dialogues is slightly higher than that of synthetic dialogues in terms of naturalness and coherence; ii)~for certain social factors, \eg\ topic, there is a higher percentage of experiment participants agreeing with the annotations of synthetic dialogues than those of human-written dialogues. 

\subsection{Ontology for {Social Factors}}
\label{sec:ontology}

{\citet{hovy-yang-2021-importance} introduced a taxonomy of social factors in order to understand what is missing in current NLP research. To study Chinese social factors using politeness theory, we focus on factors such as \textit{social norms}, \textit{social distance}, \textit{social relation} and \textit{social context}, which includes topic of conversation and location. At the core of the ontology are \textbf{\textit{norm rules}}, which are represented in a semi-structured format that incorporates the relations between social norms and other social factors, in order to facilitate research on social impact analysis in dialogues.}







Given a situation, social norms tell us what conduct is culturally appropriate. In society, social norms provide guidelines on socially acceptable or unacceptable behaviours. Consequently, social norms set implicit expectations of people's judgements and reactions to other people's conduct. Hence, a violation of social norms often leads to negative emotions, such as sadness and anger, and even disengagement or break-up of relationships. 

Social norms can be represented as rules, such as free-text rules-of-thumb (RoTs) in~\cite{forbes-etal-2020-social}. RoTs are easy for people to understand, but are challenging for machines to comprehend. 
In addition, a norm rule is often associated with other supporting social factors, such as topic and social relation, that support the characterization of a situation. Therefore, we represent a norm rule in the following semi-structured format by incorporating supporting social factors as conditions of the rules:

{\small \texttt{IF $F_a$ = k AND ... AND $F_j$ = t,\\
\indent
THEN $action$, which adheres to or violates social norms.}}
        
\noindent
where $F_i$ denotes a social factor associated with a rule, the lowercase characters are the values of those factors, and $action$ describes a behaviour in text that is considered as adhering or violating social norms. 
Next, we introduce the factors that comprise a norm rule.

\paragraph{Social Norm.} In the field of social science, social norms can be categorized by what people intend to do~\cite{cialdini1998social,bicchieri2005grammar}.
If a person intends to make a request or an apology, the expected behaviours will be clearly different. In this work, we refer to intentions as norm categories, and focus on five social norm categories due to their commonality in daily life: \emph{greeting}, \emph{request}, \emph{apology}, \emph{persuasion} and \emph{criticism}. Each of these norm categories is associated with sub-categories that provide further details about possible situations. All norm categories, sub-categories and actions are listed in Table~\ref{tab:social_norm}.

\begin{table}[!t]
    \centering
    \resizebox{.48\textwidth}{!}{
    
    \begin{tabular}{c|c|c}
    \toprule
       {\bf Social Norm}  & {\bf Sub-Category} & {\bf Action} \\ \hline
        \multirow{3}{*}{Greeting }&  business greeting & use standard greeting  \\ \cline{2-3}
              &  public greeting     &  use acceptable words for greeting         \\\cline{2-3}
              &  casual greeting    &  can use informal but polite words            \\ \hline
        \multirow{4}{*}{Request} & direct command & use a direct command\\\cline{2-3}
          &   polite command & use a politeness marker to give polite commands                  \\\cline{2-3}
              & preparatory request & should ask about preparatory conditions for a request                    \\ \cline{2-3}
              &   hint request &  request through strong hints                  \\\hline
        \multirow{3}{*}{Apology} & light apology & should give light apology\\\cline{2-3}
          &  strong apology &   should use stronger apology words                  \\\cline{2-3}
              &  party apology &  apology on behalf of the offender                \\ \hline
        \multirow{2}{*}{Persuasion} & direct persuasion & logic, comparisons of pros and cons\\\cline{2-3}
          &  indirect persuasion   & use indirect persuasion to persuade               \\ \hline
       \multirow{2}{*}{ Criticism} & indirect criticism & use indirect, subtle language \\\cline{2-3}
         &   direct criticism &    use direct criticism, strong tone and display emotions               \\
    \bottomrule
    \end{tabular}
    }
    \vspace*{2mm}
    \caption{\ourcorpus contains 5 social norm categories and 14 sub-categories. Norm violations can be avoided by following the action requirement of each social norm rule.}
    \label{tab:social_norm}
\vspace{-2mm}
\end{table}

\paragraph{Social Distance.} In sociology, social distance measures the degree of closeness or acceptance between individuals or social groups~\cite{wolfson1990bulge}. In this ontology, we broadly divide frequently observed social distance between two interlocutors into five categories: \emph{family}, \emph{friend}, \emph{romantic}, \emph{working}, and \emph{stranger}. 

\paragraph{Social Relation.} The above categories of social distance do not always reflect the power distance between two people. In real life, the power distance between two interlocutors often determines if a behaviour is culturally acceptable or not. For example, it is acceptable for an employee to greet his colleague informally, but is inappropriate to use an informal greeting for his director. Therefore, we introduce the following social relations in order to better capture relationships between interlocutors: \emph{peer-peer}, \emph{elder-junior}, \emph{chief-subordinate}, \emph{mentor-mentee}, \emph{commander-soldier}, \emph{student-professor}, \emph{customer-server} and \emph{partner-partner}.

\paragraph{Social Context.} Social norms also vary depending the detailed settings of contexts, including formality, the background of conversations and where a conversation takes place~\cite{hovy-yang-2021-importance}.
\begin{itemize}
  \item \textbf{Formality}: A situation is either \emph{formal} or \emph{informal}. A formal situation includes business meetings and conferences, while an informal context includes a chat between family members at home.
  \item \textbf{Location}: \emph{open-area}, \emph{online}, \emph{home}, \emph{police-station}, \emph{refugee campus}, \emph{restaurant}, \emph{store} and \emph{hotel}.
  \item \textbf{Topic}: We consider conversations about \emph{sales}, \emph{life-trivial}, \emph{office-affairs}, \emph{school-life}, \emph{food}, \emph{farming}, \emph{poverty-assistance}, \emph{police-corruption}, \emph{counter-terrorism} and \emph{child-missing}. {These topics were sourced from LDC~\cite{ldc2022}.}
\end{itemize}

Norm category and status (adhered or violated) are annotated at the utterance level, and the remaining factors (social distance, social relation and social context) are annotated at the dialogue level. We show norm category annotations at the dialogue level if a dialogue is associated with only one norm category. 

\subsection{Curating the Evaluation Data Via Crowd-Sourcing}
\label{sec:test_data}

\begin{figure}
    \centering
    \includegraphics[width=0.47\textwidth]{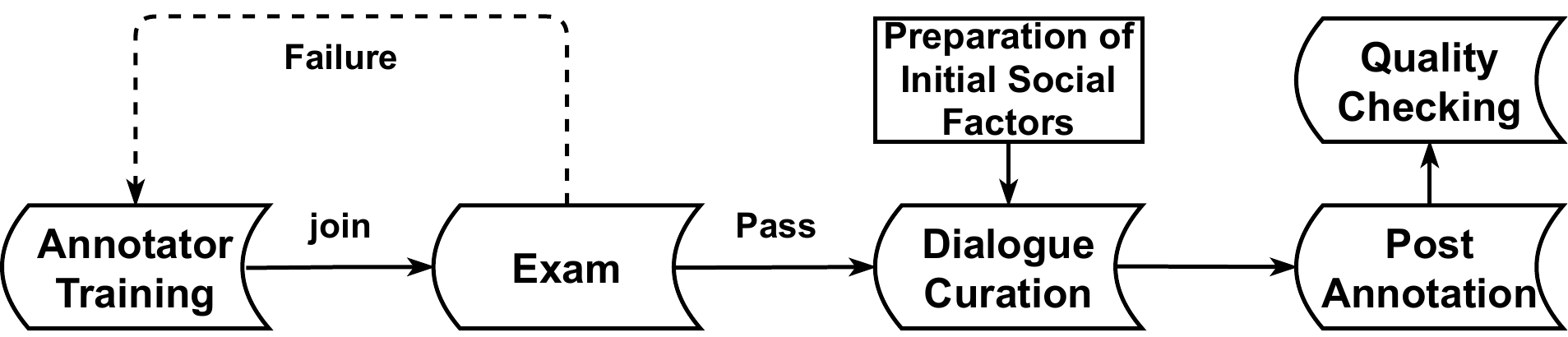}
    \caption{Process of the curation of evaluation data via crowd-sourcing.}
    \label{fig:human_curation}
\vspace{-1mm}
\end{figure}

{To facilitate our research of socially-aware dialogue systems, we  first collected a set of socially-aware conversations via crowd-sourcing.} 
\footnote{{We obtained ethics approval for our human data collection protocol from the Monash University Human Research Ethics Committee (project ID: 32962).  
}} As high-quality conversation examples are required for this complex task, we took great care to ensure the effectiveness of social factors in conversations. Specifically, (1)~as our socially-aware dialogue corpus is based on Chinese social norms, the crowd-workers should be conversant with Chinese culture --- we therefore invited 10 Chinese graduate students for the dialogue crowd-sourcing process; and (2)~we devised and used multiple human and automatic mechanisms to check the quality of collected raw dialogue data and annotations. The overall process for the curation of the test data via crowd-sourcing is presented in Figure~\ref{fig:human_curation}.

\paragraph{Training with Guidelines}
We recruited {ten} university students as crowd-workers for writing dialogues annotated with labels pertaining to social factors and norms. To ensure that crowd-workers provide effective socially-aware dialogues annotated with appropriate social factors, we designed a training tutorial. In our tutorial, we presented a detailed definition of each social factor, with two or three sample conversations per factor. In the final stage of the training, crowd workers were required to take an exam where we asked them to create one dialogue for each norm category by following the examples in the the tutorial. For each dialogue, they had to consider all related social factors (\eg\ topic, relation and location). Crowd-workers proceeded to the dialogue curation stage only if they passed the exam. Otherwise, they had to re-do the training. 

\paragraph{Preparation of Initial Social Factors} To encourage crowd-workers to incorporate social factors during dialogue curation, and still have enough latitude to creatively write new dialogues, we prepared an initial set of social factors for each potential dialogue, which covers \textit{norm category}, \textit{action} and \textit{topic}, which serve as scenario constraints. 
The remaining factors are left to the post-annotation stage, described below.

\paragraph{Dialogue Curation}
In the next stage, we instructed each crowd-worker to write a dialogue for each provided initial set of social factors. When writing a dialogue, crowd-workers were required to consider the following constraints: (1)~each dialogue should contain all the social factors in the initial set;
(2)~{for each dialogue, there should be at least one utterance that violates or adheres to the norm rule,} 
\eg\ as seen in Figure~\ref{fig:intro}, the second utterance violates the rule of norm category \textit{request}; and (3)~each dialogue should contain at least four utterances.

\paragraph{Post Annotation}
In the post annotation stage, we asked crowd-workers to complete the annotations for the remaining social factors, including \emph{status}, \emph{location}, \emph{social distance} and \emph{social relation}. As a result, each dialogue was annotated with a complete set of social factors. Note that all social factors, except \emph{norm status}, are annotated at the dialogue level.  

In total, we collected 1563 human-written dialogues via crowd-sourcing, {with an average of 156.3 dialogues per crowd-source worker.}

\begin{figure}[t]
    \centering  
    \includegraphics[width=0.48\textwidth]{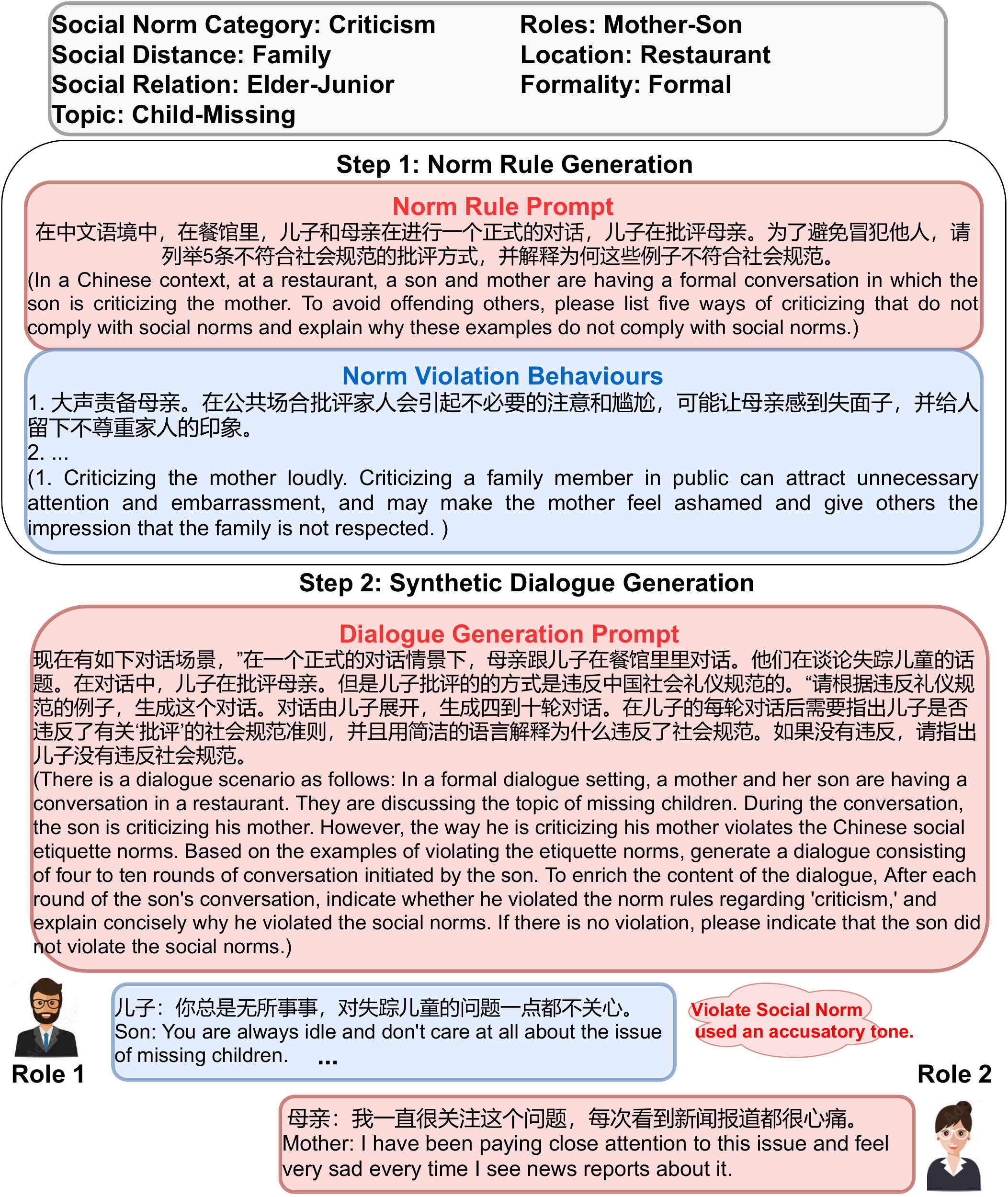}
    \caption{Process of curating synthetic data through ChatGPT -- sample input and one generated dialogue.}
    \label{fig:chatGPT}
\vspace{-1mm}
\end{figure}

\subsection{Automated Curation of Synthetic  Data}
\label{sec:training_data}
Collecting large-scale dialogues with social knowledge could be time-consuming and costly. Therefore, we propose to prompt ChatGPT to generate synthetic dialogue data automatically. As shown in Figure~\ref{fig:chatGPT}, the main steps for generating annotated synthetic dialogues are: (1)~\textit{Norm Rule Generation}, and (2)~\textit{Synthetic Dialogue Generation}. The prompts for both steps are constructed by entering values of social-factor variables into template slots. Clearly, the cost of generating synthetic data is much lower than that of collecting human-written data. 
\paragraph{Norm Rule Generation} In this step, we provide ChatGPT with \textit{Norm Rule Prompts} generated by incorporating combinations of roles, formalities, social norm categories and locations into a template. 
We offer more than ten role pairs within the prompts to ensure a wide variety of social relationships and distances. 
For example, the ``mother'' and ``son'' roles implicitly convey the social distance of ``family'' and the relationship between ``elder-junior''.
However, instead of directly integrating abstract social relationships and distances into the templates, we use a natural language description to incorporate these factors in the prompt context.
Each prompt directs ChatGPT to generate a series of behaviours that violate social norms in a specific context. As depicted in Figure~\ref{fig:chatGPT}, we requested ChatGPT to identify various examples of norm-violating behaviours that could occur in a restaurant during a formal conversation between a son (role 1) and mother (role 2), where the son is criticizing the mother (norm violation).


\paragraph{Synthetic Dialogue Generation}
In this phase, we present ChatGPT with \textit{Dialogue Generation Prompt}, augmented by prefixing to it the Norm Rule Prompt and generated norm-violating behaviours. The Dialogue Generation Prompt incorporates \textit{original labels} that represent norm-related social factors used in Norm Rule Prompt and a new {topic, which is randomly chosen from the topic category,}
to increase content diversity within the generated dialogues.
For instance, we prompt ChatGPT to create a dialogue about child-missing (topic) in a restaurant (location) where a son (role 1) and his mother (role 2) are having a formal (formality) conversation, during which the son criticizes (norm category) his mother in a norm-violating manner. By including topics in the dialogue prompt, we can simulate a more realistic social context, enabling ChatGPT to generate nuanced and contextually relevant dialogues. In addition, we ask ChatGPT to identify the specific utterances in the dialogue that violate criticism norms for our downstream task, norm violation detection. Ideally, with the guidance from norm-violating behaviours generated in Step~1, ChatGPT should be able to more accurately identify instances of norm-violating utterances.

\subsection{Data Quality Check}
\label{sec:quality_check}

\paragraph{Procedure}
We recruited 16 university students, who are either Chinese or understand Chinese culture well, to check the quality of a sample of the
generated dialogues and their associated original labels. Henceforth, these students are called \textit{annotators}. 
The annotators were asked to attend a briefing that explained the definitions of the social factors and the task. 
All the incorrect answers were explained to the annotators before they started working on the task, and people who failed were re-trained until they passed the test.

In total, 300 dialogues were checked. These dialogues, which were randomly sampled from the collected dialogue corpora, comprise
100 human-written dialogues and 200 synthetic dialogues. Each annotator was given 28-29 dialogues that are a (randomly shuffled) mix of both types of dialogues.
The annotators were given a questionnaire comprising statements about the quality of their allocated dialogues and their associated original labels; they were
asked to rate each statement on a three-point scale (\emph{agree}, \emph{unsure} and \emph{disagree}, with scores 1, 0 and -1 respectively). All annotators were asked to provide feedback if they were unsure or disagreed with a statement.   

To compute inter-annotator agreement (IAA), 139 out of the 300 dialogues were validated by at least two annotators --- these dialogues were proportionally sampled ($1/3$ human-generated and $2/3$ ChatGPT-generated). We chose Cohen’s Kappa score~\cite{cohen1960coefficient} as the IAA metric. The Kappa score for the 139
dialogues is 0.540 (0.541 for human-written and 0.539 for synthetic dialogues).  


\begin{figure*}[t]
    \centering
     \subfigure[Naturalness and coherence]{
        \begin{minipage}[t]{0.30\textwidth}
        \centering
        \includegraphics[width=1\textwidth]{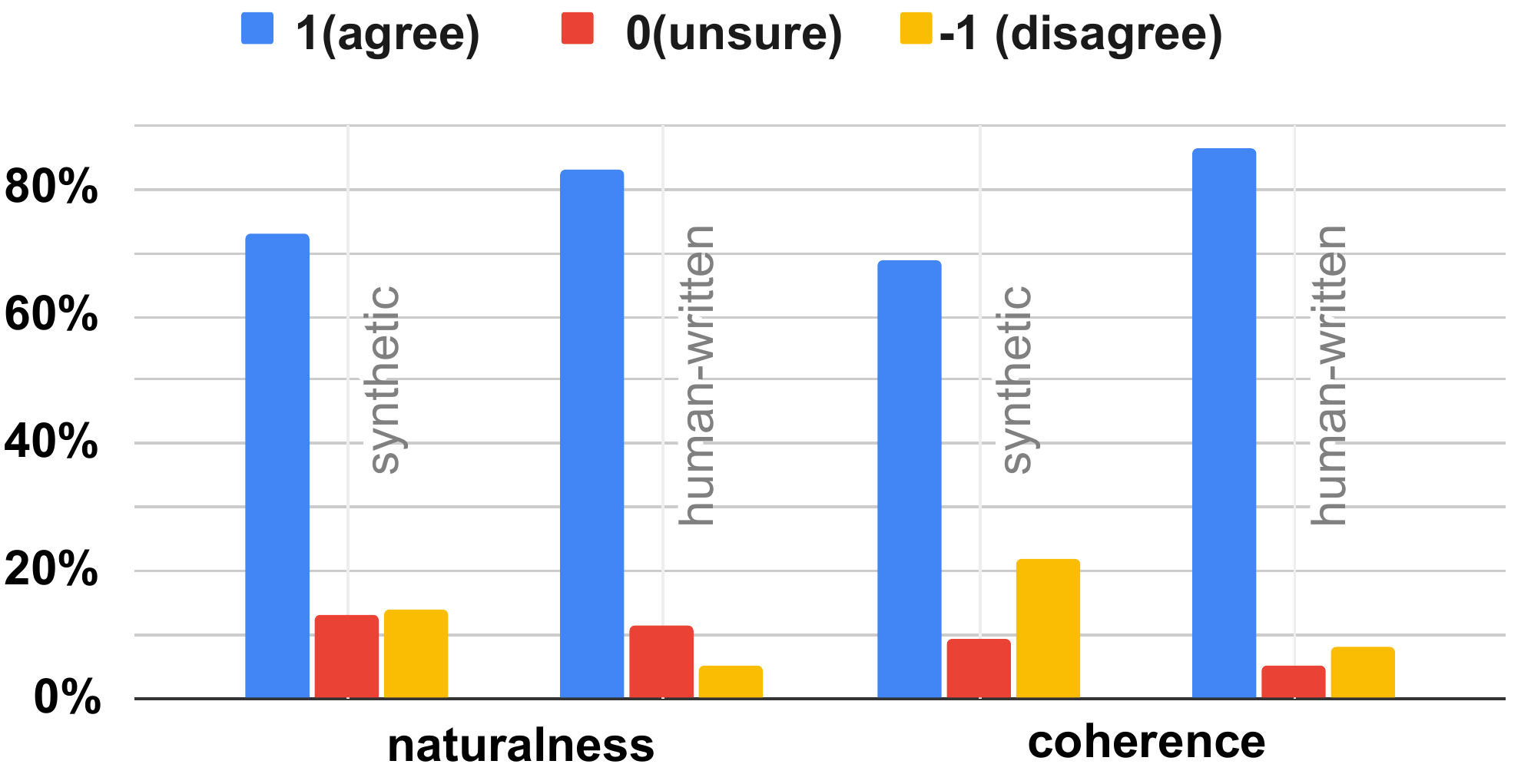}
        \label{fig:natural_coherence}
        \end{minipage}
    }
    \subfigure[Ratios of \textit{Unsure} ratings for different social factors]{
        \begin{minipage}[t]{0.31\textwidth}
        \centering
        \includegraphics[width=1\textwidth]{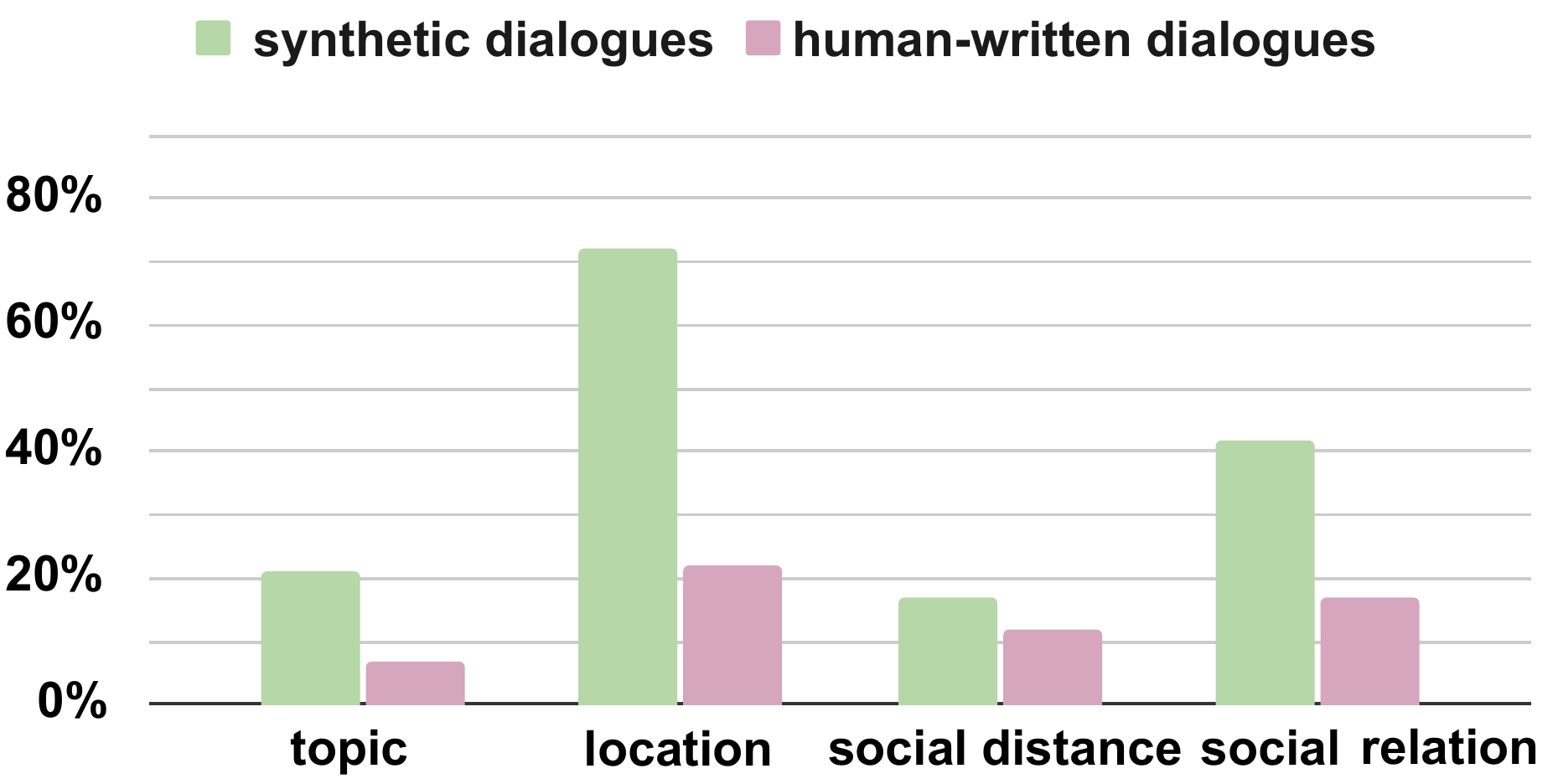}
        \label{fig:unsure}
        \end{minipage}
    }
    \subfigure[Ratings for norm rules]{
        \begin{minipage}[t]{0.295\textwidth}
        \centering
        \includegraphics[width=1\textwidth]{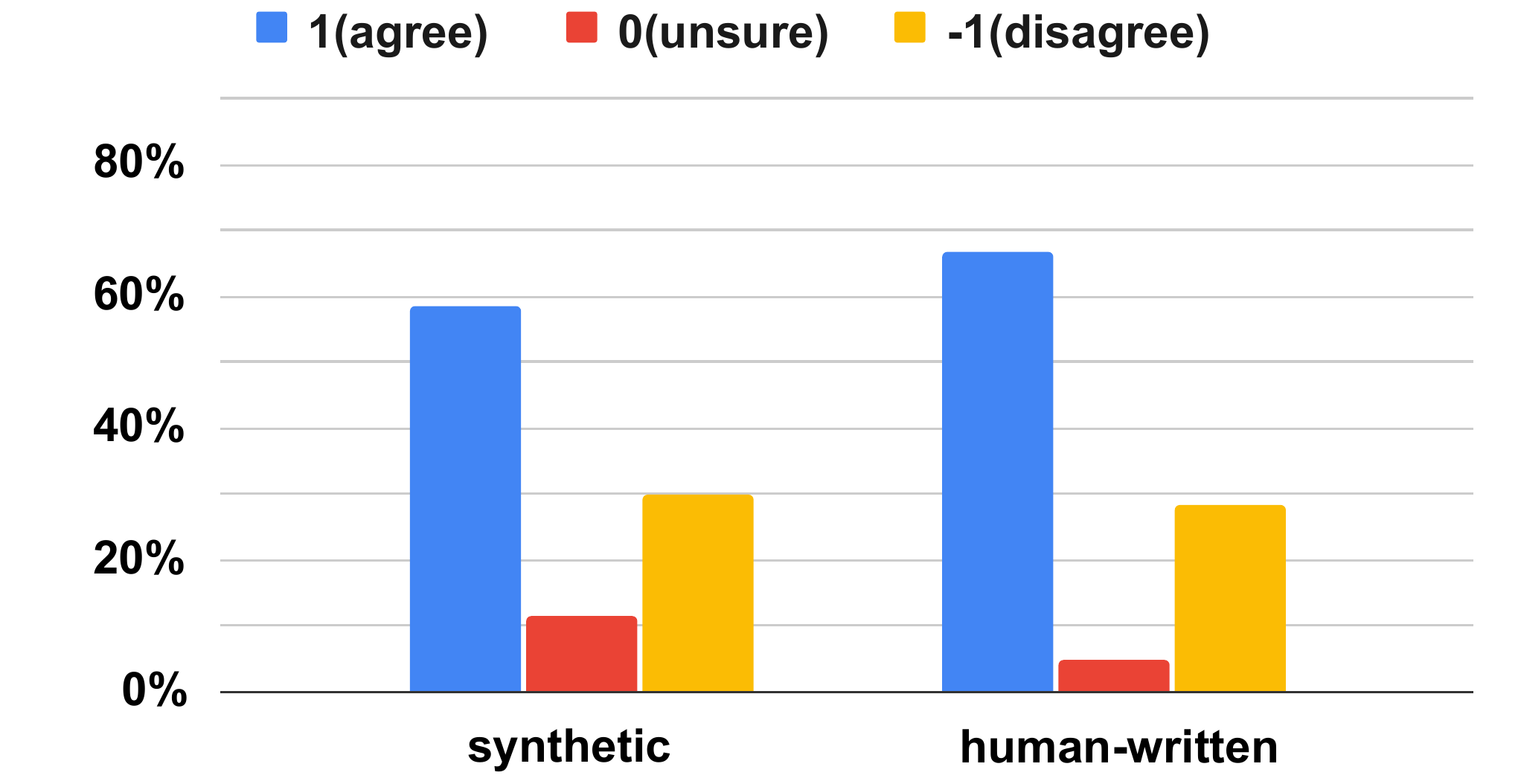}
        \label{fig:rule}
        \end{minipage}
    }
    \vspace*{-2mm}
    \caption{Annotation quality check for human-written and synthetic dialogues: (a)~naturalness and coherence, (b)~social factors, (c)~norm rules.}
    \label{fig:unsure_rule}
\end{figure*}

\begin{figure*}[!t]
    \centering
    \subfigure[Human-written dialogues]{
        \begin{minipage}[t]{0.45\textwidth}
        \centering
        \includegraphics[width=1\textwidth]{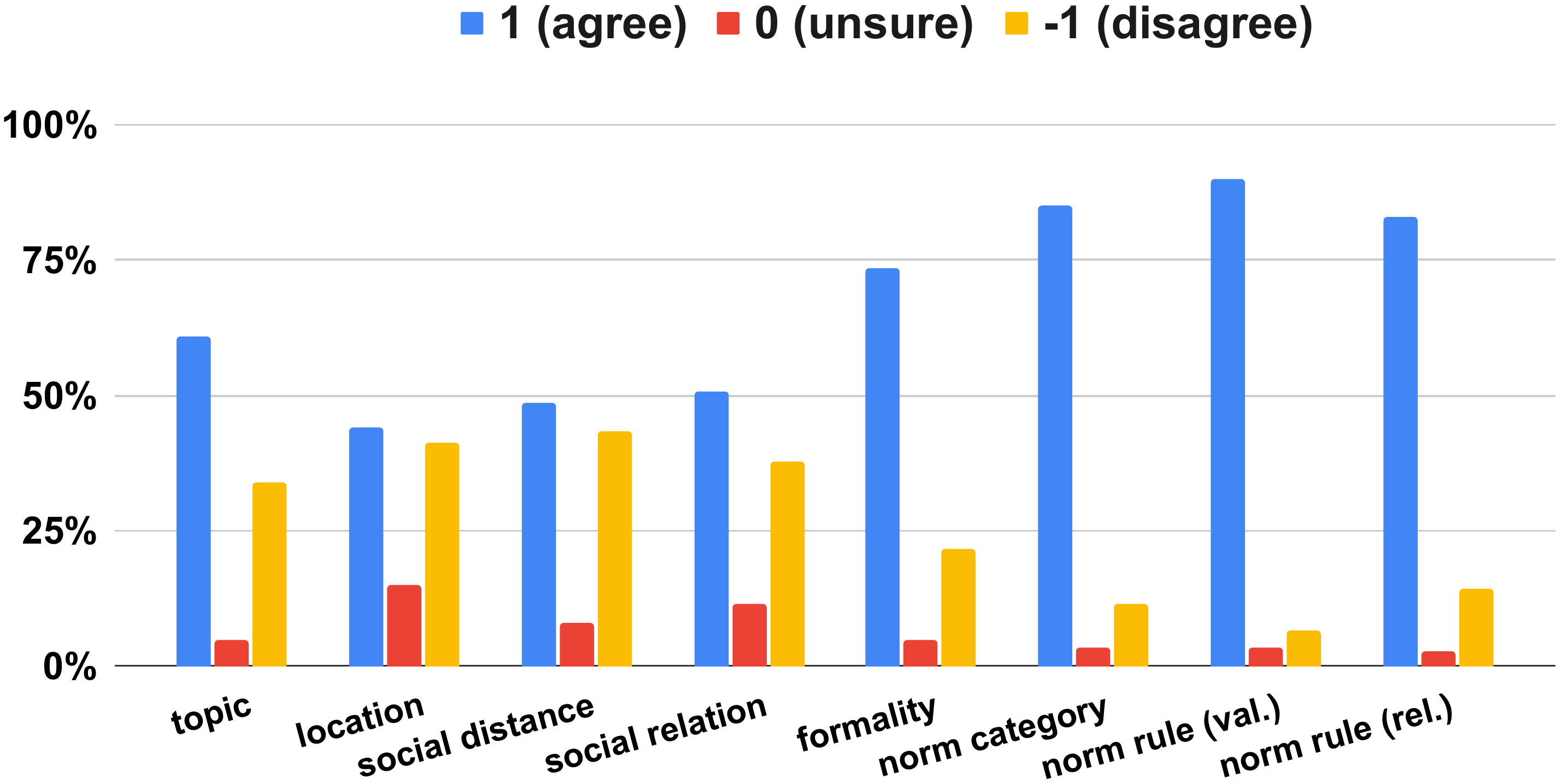}
        \label{fig:self_written}
        \end{minipage}
    }
    \subfigure[Synthetic dialogues]{
        \begin{minipage}[t]{0.45\textwidth}
        \centering
        \includegraphics[width=1\textwidth]{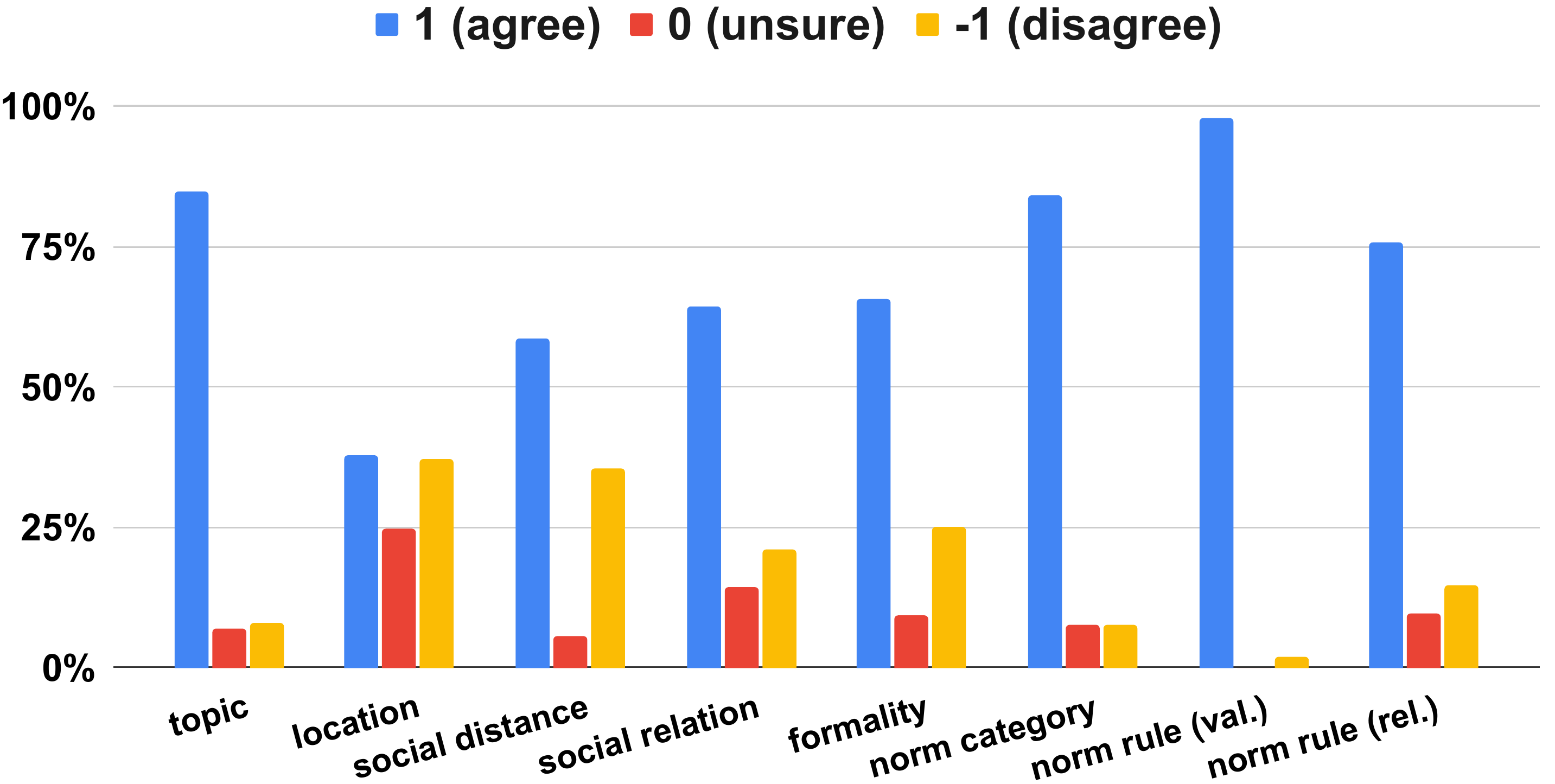}\
        \label{fig:synthetic}
        \end{minipage}
    }
    \vspace{-2mm}
    \caption{Ratings distributions for different social and norm-related factors in (a)~human-written dialogues and (b)~synthetic dialogues.}
    \label{fig:quality_check}
\vspace{-1mm}
\end{figure*}

\paragraph{Dialogue Quality.} In the validation questionnaires, the annotators were asked to assess the {naturalness} and {coherence} of each {human-written} and {synthetic} dialogue 
-- possible choices are \textit{agree}, \textit{unsure} or \textit{disagree}.
Figure~\ref{fig:natural_coherence} displays the results for each type of dialogue, which indicate that over 70\% of the annotators agree that both the human-written and the synthetic dialogues are natural and coherent.
To understand the reasons why certain dialogues were not deemed natural or coherent, we investigated the annotators' feedback. Most of the feedback mentioned that the synthetic dialogues contain some grammatical errors.

\paragraph{Annotation Quality.} The questionnaire includes questions assessing the quality of the original label of each social factor -- possible answers are \textit{agree}, \textit{unsure} and \textit{disagree}.
Even though ChatGPT and crowd workers were instructed to produce dialogues with specific original labels, as seen in Figure~\ref{fig:unsure}, locations and social relations were often not clearly identifiable in the ChatGPT-generated dialogues,
prompting annotators to give them an \textit{unsure} rating. This indicates that more careful prompt engineering is warranted.
Figure~\ref{fig:rule} shows that the annotators reach a slightly higher agreement with the original labels of the human-written dialogues than those of the synthetic dialogues.

Figure~\ref{fig:quality_check} shows the ratings of the quality-checking team with respect to the original labels of the social and norm-related factors for human-written and synthetic dialogues (Figures~\ref{fig:self_written} and~\ref{fig:synthetic} respectively). 
Here, we are concerned with (i)~if the rules are valid in Chinese culture {(norm rule (val.)); (ii)~if the annotated norm 
categories and the rules are relevant to the corresponding utterances (norm rule (rel.)); and (iii)~whether the status annotations are correct. 
Overall, there some differences in the proportion of agreement with respect to some factors, in particular topic, between the synthetic dialogues
and the human-written dialogues. In addition, for both types of dialogues,
the norm-related factors had higher ratings than most social factors (only topic in synthetic dialogues received a rating commensurate with that of the norm-related factors).

\begin{table*}[t]
\begin{adjustbox}{ width=\textwidth,center}
\centering
\begin{tabular}{lccccccc}
\hline
\textbf{Dataset}   & \textbf{Type} &  \textbf{\#Dialogues} & \textbf{\#Avg. turns} & \textbf{language} & \textbf{social factors} & \textbf{Mainly targeted tasks} & \textbf{Latest Updates }\\
\hline
\FactAct~\cite{dutt-etal-2020-keeping} & multi-turn &   299 & 35.8  & English & persuasion & face act prediction & 2018 \\
\PersuasionforGood~\cite{wang2019persuasion} & multi-turn & 1017 & 10.43  & English & request, persuasion & persuasion strategy prediction & 2019\\
CPED~\cite{chen2022cped} & multi-turn & 12k & 11.08 & Chinese & emotion & emotion classification & 2022  \\
\moralInt~\cite{ziems2022moralIntegrityCorpus} & single-turn &  38k & - & English & norm rule & norm classification & 2022 \\
DREAM~\cite{gu2022dream} & single-turn & 49k & - & English & norm rule & norm classification  & 2022 \\ \hline
\ourcorpus - \textit{Human-Written}  & multi-turn  & 1563 & 6.63 &  & \multirow{1}{*}{1. norm rule, 2. emotion} &  & \\ 
\ourcorpus - \textit{Synthetic}  &  multi-turn & 4870 & 10.67 & Chinese & { 3. social distance, 4.social relation}  & {classification on multiple}  & 2023\\ 
 \ourcorpus - \textit{total}  & multi-turn   & 6433 & 9.45 &   & {5.location, 6. formality} & {social factors} & \\ 

\bottomrule
\end{tabular}
\vspace*{2mm}
\end{adjustbox}
\caption{Comparison between \ourcorpus and related dialogue corpora.}
\label{tab:compare}
\end{table*}

\begin{table}[!t]
    \centering
    \small
    \begin{tabular}{clrrrr}
    \toprule
       &   & \multicolumn{2}{c}{Human-written} & \multicolumn{2}{c}{Synthetic} \\ \cline{3-4}\cline{5-6}
       & Category  & Num & Proportion  & Num & Proportion \\ \hline
       \multirow{5}*{\rotatebox{90}{\textbf{social norm}}} & greetings & 296 &  18.93\% & 974 & 20.00\% \\
        & request & 440 & 28.15\% & 1151 &23.63\% \\
    & apology & 241 &  15.42\% &  651 & 13.37\% \\
          & persuasion & 278 & 17.78\% & 805 & 16.53\%\\
           & criticism & 308 &  19.70\% & 1289 & 26.47\%\\ \cline{3-6}
        & \textit{overall} & 1563 & 100\% & 4870 & 100\% \\ \cline{2-6}
           \multirow{5}*{\rotatebox{90}{\textbf{social distance}}} & family & 131 & 8.38\% & 449 & 9.22\%\\
        & friend & 257 & 16.44\% & 762 & 15.65\%\\
         & romantic & 35 & 1.56\% & 501 & 10.29\%\\
          & working & 640 &  40.94\% & 2888 & 59.30\%\\
           & stranger & 500 & 31.98\% & 270 & 5.54\%\\ \cline{3-6}
        & \textit{overall} & 1563 & 100\% & 4870 & 100\%\\ \cline{2-6}
        \multirow{8}*{\rotatebox{90}{\textbf{social relation}}} & peer-peer & 563 & 36.02\% & 1089 & 22.36\% \\
        & elder-junior & 218 &  13.95\% & 449 & 9.22\%\\
         & chief-subordinate & 431 & 27.58\% & 1383 & 28.40\% \\
          & commander-soldier & 27 & 1.73\% & 465 &9.55\% \\
           & student-professor & 102 & 6.53\% & 693 &14.23\%\\
           & partner-partner & 39 & 2.50\% & 501 &10.29\%\\
          & customer-server  & 175 & 11.20\%& 270 &5.54\%\\
           & mentor-mentee & 8 &0.51\% & 20 &0.41\%\\\cline{3-6}
        & \textit{overall} & 1563 & 100\% & 4870 & 100\%\\ \cline{2-6}
        \multirow{10}*{\rotatebox{90}{\textbf{location}}} & open-area & 301 & 19.26\%& 219 &4.50\%\\
         & online & 15 & 0.96\%& 377 &7.74\%\\
          & office & 619 & 39.60\%& 635 &13.04\%\\
           & home & 163 &10.43\% & 928 &19.06\%\\
            & police-station & 38 & 2.43\%& 278 &5.71\%\\
             & refugee-camp & 9 &0.58\% & 261 &5.36\%\\
              & school & 202 &12.92\% & 411 &8.44\%\\
               & restaurant & 94 & 6.01\%& 972 &19.96\%\\
               & store & 113 & 7.23\%& 316 &6.49\%\\
               & hotel & 9 & 0.58\%& 473 &9.71\%\\
               \cline{3-6}
        & \textit{overall} & 1563 & 100\% &4870 & 100\%\\ \cline{2-6}
        \multirow{2}*{\rotatebox{90}{\textbf{FM}}} & formal & 561 & 35.89\%& 4295 &88.19\%\\
         & informal & 1002 & 64.11\%& 575 &11.81\%\\\cline{3-6}
        & \textit{overall} & 1563 & 100\% & 4870 & 100\%\\
        
    \bottomrule
    \end{tabular}
    \vspace*{2mm}
    \caption{Statistics of the \ourcorpus dataset. ``FM'' refers to the social factor \emph{formality}.}
    \label{tab:statistics}
\vspace{-2mm}
\end{table}

\subsection{Data Statistics}
\label{sec:data_stat}
\paragraph{Comparison with other datasets}
We compare \ourcorpus with dialogue datasets involving social factors. As shown in Table~\ref{tab:compare}, 
\FactAct~\cite{dutt-etal-2020-keeping} and \PersuasionforGood~\cite{wang2019persuasion} cover persuasion strategies, which are considered a social norm category by \citet{etzioni2000social}.
To facilitate the research on socially-aware dialogues, \moralInt~\cite{ziems2022moralIntegrityCorpus} and DREAM~\cite{gu2022dream} offer large-scale socially-aware datasets by crawling websites 
{or extracting related sources from those datasets in other tracks}
(\eg\ ethics,  moral stories). However, these datasets have a simple QA style format. This restricts their application in real scenarios, which should be based on multiple interactions. 
Moreover, these datasets are based on English social norms, which differ from the social norms in Chinese culture. 
As a first step in bridging this gap, CPED~\cite{chen2022cped} provides a large-scale Chinese dataset containing over 12K dialogues with 13 different types of emotions and associated emotion classification tasks. 
\ourcorpus differs from previous datasets in the following aspects: (1)~to the best of our knowledge, it is the first socially-aware dialogue dataset based on Chinese social norms, covering five most frequent social norms;
(2)~it provides fine-grained annotations of several social factors; and (3)~in addition to a human-written dataset, we provide a {synthetic} data generation framework based on a large-scale language model, viz ChatGPT.

\paragraph{Statistics of \ourcorpus}
The overall statistics of the 6,433 dialogue examples in \ourcorpus are shown in Table~\ref{tab:statistics}, including 1,563 crowd-sourced dialogues (\emph{Human-written}) and 4,870 synthetic dialogues generated by ChatGPT (\emph{Synthetic}). As seen in Table~\ref{tab:statistics}, the distributions of annotations of each social factor between \emph{Human-written} and \emph{Synthetic} are different. For example, in the \emph{Human-written} corpus, the \emph{social relation} factor ``commander-soldier''  and the \emph{location} factors ``online'', ``refugee-camp'' and ``hotel'' are  far less frequent than the other factors; this happened because we did not give strict
instructions to the crowd workers. In contrast, the distribution of factors in the
\emph{Synthetic} corpus is more balanced.
These findings indicate that synthetic data can be used not only for data augmentation, but also for adjusting the data distribution, which can be useful for low-resource settings.





\section{Experiments}
\label{sec:experiments}
We conducted experiments to evaluate baseline performance on detecting social factors in our ontology from dialogue content, followed by an investigation of the 
effectiveness of using social factors for norm-violation detection. 

\subsection{Experimental Setup}
\subsubsection{Task Definition}
As mentioned in Section~\ref{sec:ontology}, social factors are annotated either at the dialogue level or at the utterance level, hence we define the tasks at their specific levels.

\paragraph{Dialogue Level.} Given a dialogue $\rmX = \{\vx_0, ..., \vx_l\}$, where $\vx_i$ denotes an utterance at position $i$, we consider the recognition of a social factor at the dialogue level as a $K$ ways classification task, where $K$ indicates possible values of the factor. The social factors at this level include: \textit{social distance}, \textit{social relation}, \textit{formality}, \textit{location} and \textit{topic}.

\paragraph{Utterance Level.} In practice, it is particularly interesting to understand if social norms are violated or not, so that a chatbot can provide assistance in those situations. Therefore, we formulate the utterance level tasks as five binary classification problems, each of which is concerned with whether a particular social norm category is violated or not. This task is denoted as the \textit{norm-violation detection} task.


\begin{table*}[!t]
\resizebox{\textwidth}{!}{%
\begin{tabular}{l|ccc|ccc|ccc|ccc|ccc}
\toprule
& \multicolumn{15}{c}{(a) Training on human-written Dialogue} \\
\cline{2-16}
                & \multicolumn{3}{c|}{\textbf{Social Distance}} & \multicolumn{3}{c|}{\textbf{Social Relation}} & \multicolumn{3}{c|}{\textbf{Formality}} & \multicolumn{3}{c|}{\textbf{Location}} & \multicolumn{3}{c}{\textbf{Topic}} \\ \hline
\textbf{Models} & P             & R             & F1            & P             & R             & F1            & P           & R           & F1          & P           & R          & F1          & P         & R         & F1         \\ \hline
\bert           & 68.81 (6.85)               & 64.79 (3.16)              &  65.37 (4.35)             &   53.43 (3.13)            & 55.04 (2.71)               &     53.68 (2.96)          &       86.58 (1.39)      &   87.18 (1.16)          &   86.83 (1.23)         &  48.99 (2.70)         &   48.38 (2.57)         &  48.74 (2.64)           &    38.88 (1.77)        &  40.17 (2.87)         &  39.29 (2.26)          \\
\roberta      &   84.98 (2.28)     &    73.32 (2.41)      &      76.41 (2.76)         &      53.81 (3.39)         &      55.74 (3.12)         &    54.46 (3.15)           &      87.10 (1.76)         &     87.60 (1.37)        &      87.30 (1.52)       &      49.24 (2.22)       &     49.24 (3.24)        &      48.74 (2.64)      &     46.47 (7.14)        &    42.14 (4.92)       &     43.18 (5.38)                 \\
\gpt        &     47.21 (3.74)  &   53.31 (9.23)    &   50.47 (6.97)      &   49.11 (10.67)      &  47.42 (1.50)       &    44.39 (6.32)    &  79.17 (3.28)     &   80.49 (2.87)    &   78.75 (3.85)   &  53.54 (11.34)     &   54.72 (11.32)    &   51.41 (10.77)  &     47.30 (14.31) &  53.18 (9.19)    &  44.75 (12.80)    \\ \hline\hline
& \multicolumn{15}{c}{(b) Trained on Synthetic Dialogue} \\ \cline{2-16}
 & \multicolumn{3}{c|}{\textbf{Social Distance}} & \multicolumn{3}{c|}{\textbf{Social Relation}} & \multicolumn{3}{c|}{\textbf{Formality}} & \multicolumn{3}{c|}{\textbf{Location}} & \multicolumn{3}{c}{\textbf{Topic}} \\ \hline
\textbf{Models} & P             & R             & F1            & P             & R             & F1            & P           & R           & F1          & P           & R          & F1          & P         & R         & F1         \\ \hline
\bert       & 45.88 (5.08)  & 50.38 (6.07) & 40.28 (4.31)  & 44.64 (4.10) & 43.62 (4.29) & 40.96 (4.10) & 66.66 (1.42) & 63.75 (0.90) & 55.27 (0.79) & 33.87 (4.62) & 41.26 (3.44) & 29.76 (3.17) & 43.40 (2.20) & 59.67 (7.35) & 46.39 (2.49)  \\

\roberta   & 49.65 (4.26) & 51.79 (6.03) & 42.05 (3.94) & 49.28 (4.53) & 52.42 (3.41) & 45.85 (4.46) & 61.46 (1.84)& 55.94 (1.36)& 42.32 (1.66) & 28.53 (3.67) & 36.25 (4.58) & 26.74 (3.67) & 46.20 (5.38) & 60.42 (9.56) & 47.04 (5.97)   \\
\hline\hline
& \multicolumn{15}{c}{(c) Trained on human-written Dialogue + Synthetic Dialogue} \\ \cline{2-16}
 & \multicolumn{3}{c|}{\textbf{Social Distance}} & \multicolumn{3}{c|}{\textbf{Social Relation}} & \multicolumn{3}{c|}{\textbf{Formality}} & \multicolumn{3}{c|}{\textbf{Location}} & \multicolumn{3}{c}{\textbf{Topic}} \\ \hline
\textbf{Models} & P             & R             & F1            & P             & R             & F1            & P           & R           & F1          & P           & R          & F1          & P         & R         & F1         \\ \hline
\bert       & 80.48 (4.54) & 85.79 (1.77) & 82.13 (3.78) & 75.57 (4.49) & 72.86 (3.54) & 73.24 (3.96) & 87.28 (2.04) & 88.49 (1.97) & 87.79 (2.04) & 68.36 (5.59)  & 68.25 (3.80) & 66.12 (4.25) & 86.66 (3.94) & 89.52 (5.29) & 87.13 (4.24)  \\

\roberta   & 82.33 (3.88) & 82.12 (2.00) & 81.57 (2.54) & 74.92 (3.90) & 74.03 (3.88) & 73.41 (3.50)  & 87.26 (1.43) & 88.51 (2.25) & 87.76 (1.74) & 65.58 (3.48) & 65.82 (5.74) & 64.08 (4.31) & 83.65 (5.84) & 89.25 (7.08) & 84.38 (6.20)  \\
\hline

\end{tabular}
}
\vspace*{2mm}
\caption{Experiment results of dialogue-level social factor detection where the detection models are trained on (a)~Human-Written Dialogues, (b)~Synthetic Dialogues, and (c)~Human-written Dialogues + Synthetic Dialogues.}
\label{tab:self_written_dialogue_exp}
\vspace*{-2mm}
\end{table*}

\subsubsection{Baselines}
We use the following baseline models for our experiments: (1) \bert: a BERT~\cite{devlin2018bert} model pre-trained on a large-scale Chinese corpus. (2) \roberta: a RoBERTa~\cite{liu2019roberta} model pre-trained on a large-scale Chinese corpus. (3) \gpt: the GPT-3 model~\cite{brown2020language}, which uses prompt-based methods and does not require fine-tuning. 


\subsubsection{Evaluation Details.} We split the human-written dialogues annotated with social factors into training sets (75\%) and test sets (25\%).
We report the baseline performance on the test set, after tuning hyper-parameters of the baselines on the training set. To understand the usefulness of synthetic data generated by ChatGPT, we also compare model performance trained on synthetic and human-written dialogues. As the detection of social factors is regarded as a classification task, we adopt \textit{Precision}, \textit{Recall} and \textit{F1 score} as the evaluation metrics. We conduct five-fold cross-validation to report mean values and standard deviations (in brackets).

\subsection{Results and Discussion}
\label{sec:results_discussions}
\paragraph{Performance of models trained on human-written dialogues.} 
The top segment of Table~\ref{tab:self_written_dialogue_exp} displays the results obtained by the baseline models and \gpt for detecting dialogue-level
social factors in human-written dialogues.
The results for utterance-level norm violation detection are reported in the top segment of Table \ref{tab:self_written_dialogue_utterance_exp}. 
Our results show that the language models with fewer parameters (\ie\ \bert and \roberta) perform better than the much larger model, \gpt. 
The possible reason is that \gpt adopts a zero-shot setting using no in-context knowledge of social factors, which limits its performance. 


\paragraph{Performance of models trained only on synthetic data.} Clearly, it is cheaper to acquire a large amount of synthetic data by querying ChatGPT, compared to human-written dialogues.
To investigate the usefulness of synthetic data, we train the baselines \bert and \roberta on synthetic data,
and test them on the test sets of the human-written dialogues used for cross-validation. 
As the test sets are the same, we can compare the baseline results with those trained only on human-written dialogues. The dialogue-level
results are reported in the middle segment of Table~\ref{tab:self_written_dialogue_exp}, and the utterance-level results in the middle segment of 
Table~\ref{tab:self_written_dialogue_utterance_exp}. The results show that the performance of models trained on synthetic data is significantly worse than that of
models trained on human-written data. As seen in Figure~\ref{fig:quality_check}, the annotators' agreement on social factor labels for both datasets is comparable. 
Thus, we hypothesize that the lower performance obtained for synthetic data can be attributed to distributional differences between the two datasets. 
We provide further details on this potential explanation subsequently.

\paragraph{Performance of models trained on synthetic data + human-written dialogues.} Here, we combined the synthetic data with the human-written training data 
of each fold in the five-fold cross-validation.
The results of our evaluation are presented in the bottom segments of Tables~\ref{tab:self_written_dialogue_exp} and~\ref{tab:self_written_dialogue_utterance_exp}.
They indicate that both \bert and \roberta, trained on a combination of human-written and synthetic dialogues, consistently outperform models trained 
on a single resource in terms of their ability to predict social factors and detect norm violations. These findings suggest that incorporating synthetic 
data into the training process significantly boosts the performance of models that rely on human-written dialogues.

\begin{table}[!t]
\centering
\small
\begin{tabular}{l|ccc}
\hline
  
& \multicolumn{3}{c}{(a) Trained on human-written Dialogue} \\ \cline{1-4}
\textbf{Models}                 & \textbf{P} & \textbf{R} & \textbf{F1} \\ \hline
\bert           & 46.05 (7.64)                    &   13.32 (3.45)              &   18.77 (3.70)         \\
\roberta        & 42.02 (11.34)                   &   16.55 (1.32)              &  20.70 (1.02)          \\
\gpt        &  24.61 (2.27)      &  19.77 (1.99)   & 20.56 (1.02) \\ \hline\hline 
& \multicolumn{3}{c}{(b) Trained on Synthetic Dialogue} \\ \cline{1-4}
\textbf{Models}                 & \textbf{P} & \textbf{R} & \textbf{F1} \\ \hline
\bert          & 15.23 (2.61) & 8.86 (0.96) & 10.52 (1.15)        \\
\roberta        & 14.43 (3.12) & 6.82 (0.93) & 8.80 (1.42)        \\\hline\hline 
& \multicolumn{3}{c}{(c) Trained on  Human-written + Synthetic Dialogues } \\ \cline{1-4}
\textbf{Models}                 & \textbf{P} & \textbf{R} & \textbf{F1} \\ \hline
\bert          & 63.45 (1.23) & 33.90 (1.63) & 42.86 (1.84)        \\
\roberta        & 66.83 (2.57)&33.82 (1.20) & 44.09 (1.18)        \\\hline
\end{tabular}
\vspace{2mm}
\caption{Experiment results of utterance-level social factor detection where the detection models are trained on (a)~Human-Written Dialogues, (b)~Synthetic Dialogues, and (c)~Human-written + Synthetic Dialogues.}
\label{tab:self_written_dialogue_utterance_exp}
\vspace{-2mm}
\end{table}

\paragraph{Visualization of Distribution Divergences.}
Even though the naturalness and coherence of the dialogues generated by ChatGPT are comparable with those of human-written ones, we conjecture that the
synthetic data follows different distributions from those of human-written dialogues. 
To test this conjhecture, we generate the embeddings of randomly sampled synthetic dialogues and those of randomly sampled human-written dialogues.
To ensure that the sampled dialogues do not differ due to using different values of the social factors, we start with a random sample of social factor combinations, then draw dialogues of each type matching at least one of the social factor combinations.
Herein, we apply the best-performing encoder \roberta to map each dialogue into a vector, then visualise them by using T-SNE~\cite{van2008tsne}, as demonstrated in Figure~\ref{fig:distribution_discrepancy}. 

In this figure, each synthetic dialogue is denoted as a single green dot, and each human-written dialogue is projected into a red dot.   
As seen in Figure~\ref{fig:distribution_discrepancy}, a substantial number of red dots are concentrated on the right side 
of the graph, while a small portion of red dots and green dots are mixed and scattered in the middle and left parts of the graph.
Therefore, we can clearly notice the difference in the distribution of the two types of dialogues.

We can also see that there are four clusters with clear boundaries within the graph.
We randomly sampled dozens
of data points from each cluster, and upon inspection of the attributes and 
contents of these samples, we found that within each cluster, the dialogues pertain to one topic, \eg\ counter-terrorism.
This indicates that the content of synthetic dialogues within the same topic is highly correlated and consistent.

Finally, we manually analyzed the overlapping red points and green points in Figure~\ref{fig:distribution_discrepancy}.
After comparing the attributes of the overlapping dots, we found that if two dialogues from different datasets are about the same topic, 
then regardless of the social factor combinations, these two dialogues are highly likely to be close to each other in the distribution space. 

\begin{figure}[t]
    \centering
    \includegraphics[width=0.45\textwidth]{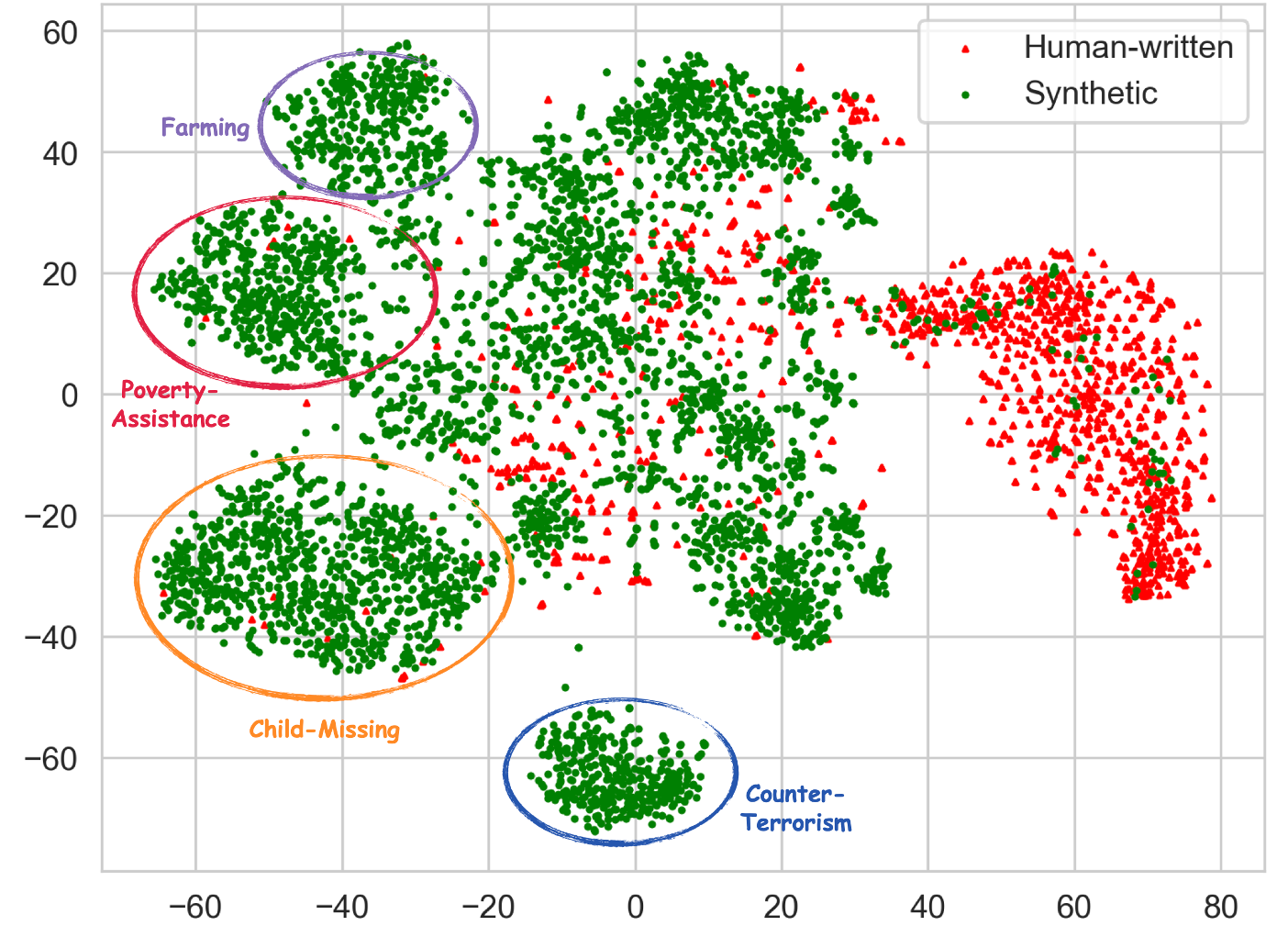}
    \caption{Distribution divergence between the embeddings of synthetic and human-written dialogues.}
    \label{fig:distribution_discrepancy}
\end{figure}

\begin{table}[!t]
\centering
\small
\resizebox{\columnwidth}{!}{%
\begin{tabular}{c|ccc}
\hline
\textbf{Models}                 & \textbf{P} & \textbf{R} & \textbf{F1} \\ \hline
\bert w/o social factors           & 46.05 (7.64)                    &   13.32 (3.45)              &   18.77 (3.70)         \\
\bert w GT social factors          & 60.31 (14.76)                   &  20.31 (1.64)               &  26.25 (2.05)           \\
\gpt w/o social factors         &  24.61 (2.27)      &  19.77 (1.99)   & 20.56 (1.02)\\
\gpt w GT social factors        &  49.99 (2.66)            &  46.96 (1.93)        &  45.53 (1.72)    \\ \hline
\end{tabular}
}
\vspace*{2mm}
\caption{Influence of social factors on the detection of social norm violations -- GT represents ground-truth social factors.}
\label{tab:application_social_norm_violation}
\vspace*{-2mm}
\end{table}

\paragraph{Incorporating social factors.} There are strong associations between social factors and social norms, including norm categories and their status. 
Therefore, we investigated if the models for predicting social-norm violations can exploit these associations to improve performance. 
Specifically, we evaluated models \bert and \gpt
by incorporating ground-truth dialogue-level social factors for the
norm violation detection task considered in Section~\ref{sec:experiments}. We added dialogue-level social factors as the context of utterances,
and fed them into the models, which were evaluated on the evaluation set of human-written dialogues.
The results in Table~\ref{tab:application_social_norm_violation} show that incorporating the social factors dramatically improves the performance of both models. 
Especially for \gpt, whose performance more than doubled. Hence, symbolic representations of social factors are indeed helpful for norm violation detection.

\section{Conclusion}
\label{section:conclusion}
In this paper, we propose \ourcorpus, a socially-aware dialogue benchmark for studying Chinese social norms in conversations, which supports
the investigation of social factors for socially-aware NLP applications. The \ourcorpus dataset contains 6,433 dialogues in total, 
of which 1,563 dialogues are written by humans via crowd-sourcing, and 4,870 synthetic dialogues are generated by ChatGPT. To the best of our knowledge, \ourcorpus is the first 
multi-turn socially-aware dialogue corpus based on Chinese social norms, annotated with social factors related to norms. 
We also offer a simple yet effective framework that utilizes large-scale language models (\eg\ ChatGPT) to generate synthetic data at scale. 
Specifically, we devised an ontology-based synthetic data generation framework, which only requires appropriate prompts for ChatGPT. 
We carefully designed the process of data collection and devised several mechanisms to ensure the quality of both dialogue curation and annotation. 

Through experiments on \ourcorpus with SOTA models, we obtained the following key findings: (1)~the quality of synthetic data is comensurate with that of 
real dialogues in terms of naturalness and coherence; (2)~SOTA model performance can be enhanced when trained on a combination of synthetic and human-written dialogues; and 
(3)~the provided social factors significantly contribute to the performance of SOTA models, such as \gpt, in social norm violation detection. 


\section*{Acknowledgements}


This material is based on research sponsored by DARPA under agreement number HR001122C0029. The U.S. Government is authorized to reproduce and distribute reprints for Governmental purposes notwithstanding any copyright notation thereon. We appreciate all annotators for their contributions to this work.

\section*{Ethics}

To mitigate possible misuse scenarios as well as  promote fairness and inclusivity, we highlight a few ethical consideration for \ourcorpus. First, we want to stress that the present resources in this work are intended for research purposes only. The social norm and violation situations in \ourcorpus should not be used as insults, slander, and other malicious intents. Besides, it is essential to recognize the mental health of crowd-workers and annotators during dataset curation process. Before data curation, this study was carefully reviewed and approved by an internal review board. We require each crowd-worker and annotator to take a rest every two hours or anytime time they do not feel comfortable.

\bibliographystyle{ACM-Reference-Format}
\bibliography{sample-base}

\end{document}